%% file: main_groupsparse.tex
\documentclass[11pt]{article}
\pdfoutput=1
\PassOptionsToPackage{numbers, compress}{natbib}
\usepackage{authblk}

\usepackage{url}

\usepackage[utf8]{inputenc} 
\usepackage{hyperref}       
\usepackage{url}            
\usepackage{booktabs}       
\usepackage{amsfonts}       
\usepackage{microtype}      

\usepackage{amssymb}
\usepackage{multirow}
\usepackage{amsmath}
\usepackage{amsthm}
\usepackage{color}
\usepackage{mathrsfs}
\usepackage{graphicx}
\usepackage{algorithmic}
\usepackage{caption, subcaption}

\usepackage{times}

\usepackage{natbib}
\usepackage[boxed,ruled]{algorithm2e}

\usepackage{comment}
\usepackage{enumitem}
\setitemize{noitemsep,topsep=0pt,parsep=0pt,partopsep=0pt}

\title{Information Projection and Approximate Inference for Structured Sparse Variables}
\author[1]{Rajiv Khanna}
\author[1]{Joydeep Ghosh}
\author[2]{Russell Poldrack}
\author[2]{Oluwasanmi Koyejo}
\affil[1]{Department of Electrical and Computer Engineering \authorcr The University of Texas at Austin \authorcr \texttt{\{rajivak, jghosh@\}@utexas.edu}}
\affil[2]{Department of Psychology \authorcr Stanford University \authorcr \texttt{\{poldrack, sanmi\}@stanford.edu}}

\begin{document} 

\newcommand{\argmin}[1]{\underset{#1}{\text{argmin}}}
\newcommand{\argmax}[1]{\underset{#1}{\text{argmax}}}
\newcommand{\ch}[1]{\text{conv}(#1)}
\newcommand{\lin}[1]{\text{lin}(#1)}
\let\chapter\section

\newcommand{\R}{\mathbb{R}}
\newcommand{\vectt}[1]{\overrightarrow{#1}}
\newcommand{\TODO}[1]{\color{red}TODO:#1.\color{black}}

\newcommand{\mt}{\mathbf{T}}
\newcommand{\mx}{\mathbf{X}}

\newcommand{\mw}{\mathbf{W}}
\newcommand{\mi}{\mathbf{I}}
\newcommand{\mvt}{\vectt{\text{T}}}
\newcommand{\mvw}{\vectt{\text{W}}}
\newcommand{\boldmu}{\boldsymbol \mu}
\newcommand{\boldepsilon}{\boldsymbol \epsilon}
\newcommand{\tildeP}{\text{Q}}

\newcommand{\PMP}{\text{P}_{\cM_\perp}}

\newcommand{\PM}{\text{P}_\cM}

\maketitle
\input{notation.tex}
\input{theorem_notation.tex}

\input{abstract.tex}
\input{introduction.tex}
\input{background.tex}
\input{groupsparse.tex}

\input{applications.tex}

\input{experiments.tex}

\input{conclusion.tex}

\newpage
{
\bibliographystyle{plainnat}
\bibliography{paper}
}
\newpage
\appendix 

\input{appendix}

\end{document}

%% file: notation.tex
\def\etal{et\/ al.\/ }
\def\bydef{:=}
\def\th{{^{th}}}
\def\suchthat{\text{s.t.}}

\def\embed{\hookrightarrow}
\def\to{{\,\rightarrow\,}}
\def\kron{\otimes}
\def\had{\circ}
\def\ksum{\oplus}

\mathchardef\mhyphen="2D
\newcommand{\smallmat}[1]{\left[\begin{smallmatrix}#1\end{smallmatrix}\right]}
\newcommand{\smallmt}[1]{\begin{smallmatrix}#1\end{smallmatrix}}

\newcommand{\tvec}[1]{{\text{vec}(#1)}}
\newcommand{\mat}[1]{{\text{mat}(#1)}} 
\newcommand{\indicator}[1]{ {\mathsf{1}}_{\left[ {#1} \right] }}

\def\half{\frac{1}{2}}
\newcommand{\fracl}[1]{\frac{1}{#1}}
\providecommand{\abs}[1]{{\left\lvert#1\right\rvert}}
\providecommand{\trans}[1]{{#1^\dagger}}
\providecommand{\inv}[1]{{#1^{-1}}}
\providecommand{\inner}[2]{{\left\langle#1, #2\right\rangle}}
\providecommand{\innere}[2]{{\trans{#1}#2}}
\providecommand{\innerd}[2]{{#1{\boldsymbol{\cdot}}#2}}
\providecommand{\brac}[1]{{\left(#1\right)}}

\providecommand{\dot}[2]{{\trans{#1}#2}}
\newcommand{\tr}[1]{{\mathrm{tr}}\!\left( #1 \right)}
\newcommand{\kl}[2]{{\mathrm{KL}\!\left({#1}\Vert{#2}\right)}}
\newcommand{\Hp}[1]{{\mathrm{H}\!\left({#1}\right)}}
\newcommand{\Hpp}[1]{-\Hp{#1}}

\newcommand{\norm}[1]{{ \left\lVert#1\right\rVert }}
\newcommand{\normb}[1]{\norm{#1}_2}
\newcommand{\normbs}[1]{\normb{#1}^2}

\newcommand{\vertiii}[1]{{\left\vert\kern-0.25ex\left\vert\kern-0.25ex\left\vert #1
    \right\vert\kern-0.25ex\right\vert\kern-0.25ex\right\vert}}

\newcommand{\normf}[1]{\vertiii{#1}_2}
\newcommand{\normfs}[1]{\normf{#1}^2}
\newcommand{\normt}[1]{\vertiii{#1}_1}
\newcommand{\normr}[1]{\vertiii{#1}_0}

\newcommand{\normh}[2]{\vertiii{#1}_{2 \mhyphen #2}}
\newcommand{\normhs}[2]{\normh{#1}{#2}^2}
\newcommand{\normht}[2]{\vertiii{#1}_{1 \mhyphen #2}}
\newcommand{\normhr}[2]{\vertiii{#1}_{0\,\mhyphen #2}}

\newcommand{\gp}[2]{\cG\cP \left(#1, #2\right)}
\newcommand{\gpv}[3]{\cM\cG\cP \left(#1, #2, #3\right)}
\newcommand{\g}[2]{\cN\left(#1, #2\right)}
\newcommand{\gv}[3]{\cM\cN \left(#1, #2, #3\right)}
\newcommand{\etr}[1]{\text{etr}\left( #1 \right)}
\newcommand{\ev}[2]{\mathrm{E}_{ #1}\left[ #2 \right]}
\newcommand{\expf}[1]{ \exp \left( #1 \right) }
\newcommand{\exf}[1]{ e^{ #1 } }
\def\sut{\text{s.t.}}

\newcommand{\D}[2]{\frac{d #1}{d #2}} 
\newcommand{\dd}[2]{\frac{d^2 #1}{d #2^2}} 
\newcommand{\pd}[2]{\frac{\partial #1}{\partial #2}}
\newcommand{\pdd}[2]{\frac{\partial^2 #1}{\partial #2^2}}
\newcommand{\gradb}[1]{\gv{\nabla} #1}
\newcommand{\grad}[1]{ \nabla_{#1} }

\newcommand{\vect}[1]{{\boldsymbol{#1}}}
\def\eps{\epsilon}
\def\balpha{\vect{\alpha}}
\def\bbeta{\vect{\beta}}
\def\bmu{\vect{\mu}}
\def\bnu{\vect{\nu}}
\def\beps{\vect{\epsilon}}
\def\bphi{\vect{\phi}}
\def\bpsi{\vect{\psi}}
\def\btheta{\vect{\theta}}
\def\bpi{\vect{\pi}}
\def\bgamma{\vect{\gamma}}
\def\beeta{\vect{\eta}}
\def\blambda{\vect{\lambda}}
\def\bkappa{\vect{\kappa}}
\def\bupsilon{\vect{\upsilon}}
\def\bomega{\vect{\omega}}
\def\btau{\vect{\tau}}
\def\brho{\vect{\rho}}
\def\bchi{\vect{\chi}}
\def\bPi{\vect{\Pi}}
\def\bPsi{\vect{\Psi}}
\def\bPhi{\vect{\Phi}}
\def\bTheta{\vect{\Theta}}
\def\bSigma{\vect{\Sigma}}
\def\bLambda{\vect{\Lambda}}
\def\bOmega{\vect{\Omega}}
\def\bDelta{\vect{\Delta}}
\def\bUpsilon{\vect{\Upsilon}}
\def\bAlpha{\vect{\vA}}
\def\bGamma{\vect{\Gamma}}
\def\bAlpha{\vect{A}}

\def\ta{{\text{a}}}
\def\tb{{\text{b}}}
\def\tc{{\text{c}}}
\def\td{{\text{d}}}
\def\te{{text{e}}}
\def\tf{{text{f}}}
\def\tg{{\text{g}}}
\def\th{{\text{h}}}
\def\ti{{\text{i}}}
\def\tk{{\text{k}}}
\def\tl{{\text{l}}}
\def\tm{{\text{m}}}
\def\tn{{\text{n}}}
\def\to{{\text{o}}}
\def\tp{{\text{p}}}
\def\tq{{\text{q}}}
\def\tr{{\text{r}}}
\def\ts{{\text{s}}}
\def\tt{{\text{t}}}
\def\tu{{\text{u}}}
\def\tv{{\text{v}}}
\def\tw{{\text{w}}}
\def\tx{{\text{x}}}
\def\ty{{\text{y}}}
\def\tz{{\text{z}}}

\def\va{{\vect{a}}}
\def\vb{{\vect{b}}}
\def\vc{{\vect{c}}}
\def\vd{{\vect{d}}}
\def\vee{{\vect{e}}}
\def\vff{{\vect{f}}}
\def\vg{{\vect{g}}}
\def\vh{{\vect{h}}}
\def\vi{{\vect{i}}}
\def\vk{{\vect{k}}}
\def\vl{{\vect{l}}}
\def\vm{{\vect{m}}}
\def\vn{{\vect{n}}}
\def\vo{{\vect{o}}}
\def\vp{{\vect{p}}}
\def\vq{{\vect{q}}}
\def\vr{{\vect{r}}}
\def\vs{{\vect{s}}}
\def\vt{{\vect{t}}}
\def\vu{{\vect{u}}}
\def\vv{{\vect{v}}}
\def\vw{{\vect{w}}}
\def\vx{{\vect{x}}}
\def\vy{{\vect{y}}}
\def\vz{{\vect{z}}}
\def\vA{{\vect{A}}}
\def\vB{{\vect{B}}}
\def\vC{{\vect{C}}}
\def\vD{{\vect{D}}}
\def\vE{{\vect{E}}}
\def\vF{{\vect{F}}}
\def\vG{{\vect{G}}}
\def\vH{{\vect{H}}}
\def\vI{{\vect{I}}}
\def\vJ{{\vect{J}}}
\def\vK{{\vect{K}}}
\def\vL{{\vect{L}}}
\def\vM{{\vect{M}}}
\def\vN{{\vect{N}}}
\def\vO{{\vect{O}}}
\def\vP{{\vect{P}}}
\def\vQ{{\vect{Q}}}
\def\vR{{\vect{R}}}
\def\vS{{\vect{S}}}
\def\vT{{\vect{T}}}
\def\vU{{\vect{U}}}
\def\vV{{\vect{V}}}
\def\vW{{\vect{W}}}
\def\vX{{\vect{X}}}
\def\vY{{\vect{Y}}}
\def\vZ{{\vect{Z}}}

\def\ba{{\mathbf{a}}}
\def\bb{{\mathbf{b}}}
\def\bc{{\mathbf{c}}}
\def\bd{{\mathbf{d}}}
\def\bee{{\mathbf{e}}}
\def\bff{{\mathbf{f}}}
\def\bg{{\mathbf{g}}}
\def\bh{{\mathbf{h}}}
\def\bi{{\mathbf{i}}}
\def\bj{{\mathbf{j}}}
\def\bk{{\mathbf{k}}}
\def\bl{{\mathbf{l}}}
\def\bm{{\mathbf{m}}}
\def\bn{{\mathbf{n}}}
\def\bo{{\mathbf{o}}}
\def\bp{{\mathbf{p}}}
\def\bq{{\mathbf{q}}}
\def\br{{\mathbf{r}}}
\def\bs{{\mathbf{s}}}
\def\bt{{\mathbf{t}}}
\def\bu{{\mathbf{u}}}
\def\bv{{\mathbf{v}}}
\def\bw{{\mathbf{w}}}
\def\bx{{\mathbf{x}}}
\def\by{{\mathbf{y}}}
\def\bz{{\mathbf{z}}}
\def\bA{{\mathbf{A}}}
\def\bB{{\mathbf{B}}}
\def\bC{{\mathbf{C}}}
\def\bD{{\mathbf{D}}}
\def\bE{{\mathbf{E}}}
\def\bF{{\mathbf{F}}}
\def\bG{{\mathbf{G}}}
\def\bH{{\mathbf{H}}}
\def\bI{{\mathbf{I}}}
\def\bJ{{\mathbf{J}}}
\def\bK{{\mathbf{K}}}
\def\bL{{\mathbf{L}}}
\def\bM{{\mathbf{M}}}
\def\bN{{\mathbf{N}}}
\def\bO{{\mathbf{O}}}
\def\bP{{\mathbf{P}}}
\def\bQ{{\mathbf{Q}}}
\def\bR{{\mathbf{R}}}
\def\bS{{\mathbf{S}}}
\def\bT{{\mathbf{T}}}
\def\bU{{\mathbf{U}}}
\def\bV{{\mathbf{V}}}
\def\bW{{\mathbf{W}}}
\def\bX{{\mathbf{X}}}
\def\bY{{\mathbf{Y}}}
\def\bZ{{\mathbf{Z}}}

\def\aa{{\vec{\ba}}}
\def\ab{{\vec{\bb}}}
\def\ac{{\vec{\bc}}}
\def\ad{{\vec{\bd}}}
\def\aee{{\vec{\be}}}
\def\aff{{\vec{\bf}}}
\def\ag{{\vec{\bg}}}
\def\ah{{\vec{\bh}}}
\def\ai{{\vec{\bi}}}
\def\ak{{\vec{\bk}}}
\def\al{{\vec{\bl}}}
\def\am{{\vec{\bm}}}
\def\an{{\vec{\bn}}}
\def\ao{{\vec{\bo}}}
\def\ap{{\vec{\bp}}}
\def\aq{{\vec{\bq}}}
\def\ar{{\vec{\br}}}
\def\as{{\vec{\bs}}}
\def\at{{\vec{\bt}}}
\def\au{{\vec{\bu}}}
\def\av{{\vec{\bv}}}
\def\aw{{\vec{\bw}}}
\def\ax{{\vec{\bx}}}
\def\ay{{\vec{\by}}}
\def\az{{\vec{\bz}}}
\def\aA{{\vec{\bA}}}
\def\aB{{\vec{\bB}}}
\def\aC{{\vec{\bC}}}
\def\aD{{\vec{\bD}}}
\def\aE{{\vec{\bE}}}
\def\aF{{\vec{\bF}}}
\def\aG{{\vec{\bG}}}
\def\aH{{\vec{\bH}}}
\def\aI{{\vec{\bI}}}
\def\aJ{{\vec{\bJ}}}
\def\aK{{\vec{\bK}}}
\def\aL{{\vec{\bL}}}
\def\aM{{\vec{\bM}}}
\def\aN{{\vec{\bN}}}
\def\aO{{\vec{\bO}}}
\def\aP{{\vec{\bP}}}
\def\aQ{{\vec{\bQ}}}
\def\aR{{\vec{\bR}}}
\def\aS{{\vec{\bS}}}
\def\aT{{\vec{\bT}}}
\def\aU{{\vec{\bU}}}
\def\aV{{\vec{\bV}}}
\def\aW{{\vec{\bW}}}
\def\aX{{\vec{\bX}}}
\def\aY{{\vec{\bY}}}
\def\aZ{{\vec{\bZ}}}

\def\bba{{\mathbb{a}}}
\def\bbb{{\mathbb{b}}}
\def\bbc{{\mathbb{c}}}
\def\bbd{{\mathbb{d}}}
\def\bbee{{\mathbb{e}}}
\def\bbff{{\mathbb{f}}}
\def\bbg{{\mathbb{g}}}
\def\bbh{{\mathbb{h}}}
\def\bbi{{\mathbb{i}}}
\def\bbj{{\mathbb{j}}}
\def\bbk{{\mathbb{k}}}
\def\bbl{{\mathbb{l}}}
\def\bbm{{\mathbb{m}}}
\def\bbn{{\mathbb{n}}}
\def\bbo{{\mathbb{o}}}
\def\bbp{{\mathbb{p}}}
\def\bbq{{\mathbb{q}}}
\def\bbr{{\mathbb{r}}}
\def\bbs{{\mathbb{s}}}
\def\bbt{{\mathbb{t}}}
\def\bbu{{\mathbb{u}}}
\def\bbv{{\mathbb{v}}}
\def\bbw{{\mathbb{w}}}
\def\bbx{{\mathbb{x}}}
\def\bby{{\mathbb{y}}}
\def\bbz{{\mathbb{z}}}
\def\bbA{{\mathbb{A}}}
\def\bbB{{\mathbb{B}}}
\def\bbC{{\mathbb{C}}}
\def\bbD{{\mathbb{D}}}
\def\bbE{{\mathbb{E}}}
\def\bbF{{\mathbb{F}}}
\def\bbG{{\mathbb{G}}}
\def\bbH{{\mathbb{H}}}
\def\bbI{{\mathbb{I}}}
\def\bbJ{{\mathbb{J}}}
\def\bbK{{\mathbb{K}}}
\def\bbL{{\mathbb{L}}}
\def\bbM{{\mathbb{M}}}
\def\bbN{{\mathbb{N}}}
\def\bbO{{\mathbb{O}}}
\def\bbP{{\mathbb{P}}}
\def\bbQ{{\mathbb{Q}}}
\def\bbR{{\mathbb{R}}}
\def\bbS{{\mathbb{S}}}
\def\bbT{{\mathbb{T}}}
\def\bbU{{\mathbb{U}}}
\def\bbV{{\mathbb{V}}}
\def\bbW{{\mathbb{W}}}
\def\bbX{{\mathbb{X}}}
\def\bbY{{\mathbb{Y}}}
\def\bbZ{{\mathbb{Z}}}
\def\cA{\mathcal{A}}
\def\cB{\mathcal{B}}
\def\cC{\mathcal{C}}
\def\cD{\mathcal{D}}
\def\cE{\mathcal{E}}
\def\cF{\mathcal{F}}
\def\cG{\mathcal{G}}
\def\cH{\mathcal{H}}
\def\cI{\mathcal{I}}
\def\cJ{\mathcal{J}}
\def\cK{\mathcal{K}}
\def\cL{\mathcal{L}}
\def\cM{\mathcal{M}}
\def\cN{\mathcal{N}}
\def\cO{\mathcal{O}}
\def\cP{\mathcal{P}}
\def\cQ{\mathcal{Q}}
\def\cR{\mathcal{R}}
\def\cS{\mathcal{S}}
\def\cT{\mathcal{T}}
\def\cU{\mathcal{U}}
\def\cV{\mathcal{V}}
\def\cW{\mathcal{W}}
\def\cX{\mathcal{X}}
\def\cY{\mathcal{Y}}
\def\cZ{\mathcal{Z}}
\def\sfa{{\mathsf{a}}}
\def\sfb{{\mathsf{b}}}
\def\sfc{{\mathsf{c}}}
\def\sfd{{\mathsf{d}}}
\def\sfee{{\mathsf{e}}}
\def\sfff{{\mathsf{f}}}
\def\sfg{{\mathsf{g}}}
\def\sfh{{\mathsf{h}}}
\def\sfi{{\mathsf{i}}}
\def\sfj{{\mathsf{j}}}
\def\sfk{{\mathsf{k}}}
\def\sfl{{\mathsf{l}}}
\def\sfm{{\mathsf{m}}}
\def\sfn{{\mathsf{n}}}
\def\sfo{{\mathsf{o}}}
\def\sfp{{\mathsf{p}}}
\def\sfq{{\mathsf{q}}}
\def\sfr{{\mathsf{r}}}
\def\sfs{{\mathsf{s}}}
\def\sft{{\mathsf{t}}}
\def\sfu{{\mathsf{u}}}
\def\sfv{{\mathsf{v}}}
\def\sfw{{\mathsf{w}}}
\def\sfx{{\mathsf{x}}}
\def\sfy{{\mathsf{y}}}
\def\sfz{{\mathsf{z}}}
\def\sfA{\mathsf{A}}
\def\sfB{\mathsf{B}}
\def\sfC{\mathsf{C}}
\def\sfD{\mathsf{D}}
\def\sfE{\mathsf{E}}
\def\sfF{\mathsf{F}}
\def\sfG{\mathsf{G}}
\def\sfH{\mathsf{H}}
\def\sfI{\mathsf{I}}
\def\sfJ{\mathsf{J}}
\def\sfK{\mathsf{K}}
\def\sfL{\mathsf{L}}
\def\sfM{\mathsf{M}}
\def\sfN{\mathsf{N}}
\def\sfO{\mathsf{O}}
\def\sfP{\mathsf{P}}
\def\sfQ{\mathsf{Q}}
\def\sfR{\mathsf{R}}
\def\sfS{\mathsf{S}}
\def\sfT{\mathsf{T}}
\def\sfU{\mathsf{U}}
\def\sfV{\mathsf{V}}
\def\sfW{\mathsf{W}}
\def\sfX{\mathsf{X}}
\def\sfY{\mathsf{Y}}
\def\sfZ{\mathsf{Z}}
\def\fra{{\mathfrak{a}}}
\def\frb{{\mathfrak{b}}}
\def\frc{{\mathfrak{c}}}
\def\frd{{\mathfrak{d}}}
\def\free{{\mathfrak{e}}}
\def\frff{{\mathfrak{f}}}
\def\frg{{\mathfrak{g}}}
\def\frh{{\mathfrak{h}}}
\def\fri{{\mathfrak{i}}}
\def\frj{{\mathfrak{j}}}
\def\frk{{\mathfrak{k}}}
\def\frl{{\mathfrak{l}}}
\def\frm{{\mathfrak{m}}}
\def\frn{{\mathfrak{n}}}
\def\fro{{\mathfrak{o}}}
\def\frp{{\mathfrak{p}}}
\def\frq{{\mathfrak{q}}}
\def\frr{{\mathfrak{r}}}
\def\frs{{\mathfrak{s}}}
\def\frt{{\mathfrak{t}}}
\def\fru{{\mathfrak{u}}}
\def\frv{{\mathfrak{v}}}
\def\frw{{\mathfrak{w}}}
\def\frx{{\mathfrak{x}}}
\def\fry{{\mathfrak{y}}}
\def\frz{{\mathfrak{z}}}
\def\frA{\mathfrak{A}}
\def\frB{\mathfrak{B}}
\def\frC{\mathfrak{C}}
\def\frD{\mathfrak{D}}
\def\frE{\mathfrak{E}}
\def\frF{\mathfrak{F}}
\def\frG{\mathfrak{G}}
\def\frH{\mathfrak{H}}
\def\frI{\mathfrak{I}}
\def\frJ{\mathfrak{J}}
\def\frK{\mathfrak{K}}
\def\frL{\mathfrak{L}}
\def\frM{\mathfrak{M}}
\def\frN{\mathfrak{N}}
\def\frO{\mathfrak{O}}
\def\frP{\mathfrak{P}}
\def\frQ{\mathfrak{Q}}
\def\frR{\mathfrak{R}}
\def\frS{\mathfrak{S}}
\def\frT{\mathfrak{T}}
\def\frU{\mathfrak{U}}
\def\frV{\mathfrak{V}}
\def\frW{\mathfrak{W}}
\def\frX{\mathfrak{X}}
\def\frY{\mathfrak{Y}}
\def\frZ{\mathfrak{Z}}

%% file: theorem_notation.tex
\newtheorem{theorem}{Theorem}
\newtheorem{definition}[theorem]{Definition}
\newtheorem{corollary}[theorem]{Corollary}
\newtheorem{lemma}[theorem]{Lemma}
\newtheorem{proposition}[theorem]{Proposition}
\renewcommand{\text}[1]{{\textnormal{#1}}}

\newcommand{\figref}[1]{{Fig.}~\ref{#1}}
\newcommand{\secref}[1]{{Section}~\ref{#1}}
\newcommand{\Secref}[1]{{Section}~\ref{#1}}
\newcommand{\tabref}[1]{{Table}~\ref{#1}}
\newcommand{\thref}[1]{{Theorem}~\ref{#1}}
\newcommand{\algref}[1]{{Algorithm}~\ref{#1}}
\newcommand{\aref}[1]{{Algorithm}~\ref{#1}}
\newcommand{\propref}[1]{{Proposition}~\ref{#1}}
\newcommand{\lemref}[1]{{Lemma}~\ref{#1}}
\newcommand{\corref}[1]{{Corollary}~\ref{#1}}
\newcommand{\chref}[1]{{Chapter}~\ref{#1}}
\newcommand{\defref}[1]{{Definiton}~\ref{#1}}

%% file: abstract.tex
\begin{abstract}
Approximate inference via information projection has been recently introduced as a general-purpose approach  for efficient probabilistic inference given sparse variables. This manuscript goes beyond classical sparsity by proposing efficient algorithms for approximate inference via information projection that are applicable to any structure on the set of variables that admits enumeration using a \emph{matroid}. We show that the resulting information projection can be reduced to combinatorial submodular optimization subject to matroid constraints. Further, leveraging recent advances in submodular optimization, we provide an efficient greedy algorithm with strong optimization-theoretic guarantees. The class of probabilistic models that can be expressed in this way is quite broad and, as we show, includes group sparse regression, group sparse principal components analysis and sparse canonical correlation analysis, among others. Moreover, empirical results on simulated data and high dimensional neuroimaging data  highlight the superior performance of the information projection approach as compared to established baselines for a range of probabilistic models.
\end{abstract}

%% file: introduction.tex
\section{Introduction}

Parsimonious Bayesian models are being increasingly used for improving both robustness and generalization performance in applications involve large amounts of data and variables. They are especially well-suited to incorporating domain knowledge by attuning the prior design to apriori knowledge and constraints at hand. For instance, sparsity constraints and associated models have gained eminence in several fields where apriori knowledge corresponding to sparsity constraints may be incorporated via the use of sparsity inducing priors. 

A natural extension to the classical notion of sparsity is \emph{structured} sparsity -- where the sparse selection of variable dimensions includes additional information. Some examples of structured sparsity include \emph{smoothness}~\citep{koyejo2014,khanna2015}, group sparsity~\citep{Witten2009, jenatton2010sspca,Liu10slep,Simon13asparse-group}, tree/graph sparsity~\citep{hegde2015} and so on. While there is a significant body of literature on classically sparse probabilistic models, including~\citep{Archambeau2009,koyejo2014,wipf2007,sheikh12,khanna2015} probabilistic models for structured sparsity have been far less studied. Our work seeks to bridge this gap for a large family of information projection based techniques.

The information projection of a distribution to a constraint set is given by the argument that minimizes the Kullback-Leibler (KL) divergence while satisfying the constraints. The use of information projection for probabilistic inference with structured variables was recently proposed by~\citet{koyejo2014}.
While information projection is a general approach, its application requires the design of efficient algorithms that are specific to pre-specified structural constraints of interest. \citet{koyejo2014} focused on the case of sparsity, where the constraint structure is given by the union of sparse supports. For this case, they proposed approximate inference by information projection to the support that captures the largest probability mass. They showed that the resulting KL minimization can be reduced to a combinatorial submodular optimization problem, and then applied a greedy algorithm for efficient approximate inference. 
 Subsequently, a similar mechanism was developed for sparse principal components analysis (sparse PCA)~\citep{khanna2015}. 

This manuscript goes beyond sparsity by proposing efficient algorithms for approximate inference via information projection that are applicable to any {\em structured sparse} variable settings which admits enumeration using a \emph{matroid}. The class of probabilistic models that can be expressed in this way is quite broad, and as we show, includes group sparse regression, group sparse principal components analysis and sparse canonical correlation analysis, among others. 
While, the original algorithm 
for the classic case of sparse variables is recovered as a special case, the generalized framework introduced in this paper is not a simple extension, but rather involves new techniques and results.

Specifically, our main contributions are as follows: 
\begin{itemize}
\item we present a framework for approximate inference via information projection for any constraint that can be enumerated as a matroid.
\item we present an efficient scheme for this inference using a greedy algorithm. 
For general matroids, an approximation of $1/2$ to the best possible approximation is guaranteed. However, for some special cases such as cardinality constraints (classical sparsity), and group sparsity, stronger guarantees of $1-1/e$ are available.
\item we show that the special cases of information projection under group sparsity and multi-view sparsity are submodular with knapsack constraint and partition matroid constraints respectively. These constraints are applied to develop new algorithms for sparse principal components analysis (PCA), and sparse canonical correlation analysis (CCA).
\item we present empirical results on high dimensional neuroimaging data that highlight the performance of the information projection approach as compared to established baselines for a range of probabilistic models.
\end{itemize}
We also apply our framework to group-sparse regression. Its development and experiments on simulated data are presented in the supplement. 


%% file: background.tex
\section{Notation and Background}
\label{sec:background}

We begin by outlining some notation. We represent vectors as small letter bolds e.g. $\bu$. Matrices are represented by capital bolds e.g. $\bX, \bT$.  Matrix transposes are represented by superscript ${\cdot}^\top$. Identity matrices of size $s$ are represented by $\mi_{s}$. $\mathbf{1} (\mathbf{0})$ is a column vector of all ones (zeroes). The $i^\text{th}$ row of a matrix $\mathbf{M}$ is indexed as $\textbf{M}_{i, \cdot}$, while $j^\text{th}$ column is $\textbf{M}_{\cdot,j}$. We use $\text{p}(\cdot), \text{q}(\cdot)$ to represent probability densities over random variables which may be scalar, vector, or matrix valued which shall be clear from context. Sets are represented by sans serif fonts e.g. $\sfS$, complement of a set $\sfS$ is $\sfS^c$. For a vector $\mathbf{u} \in \R^d$, and a set $\sfS$ of support dimensions with $|\sfS|=k, k\le d$, $\mathbf{u}_\sfS \in \R^k$ denotes subvector of $\mathbf{u}$ supported on $\sfS$. Similarly, for a matrix $\mx \in \R^{n\times d}$, $\mx_{\sfS} \in \R^{k\times k}$ denotes the submatrix supported on $\sfS$. We denote $\{1,2,\ldots, d\} $ as  $[d]$.  Let $\frp(d)$ be the power set of $[d]$. 

\textbf{Relative Entropy:} Let $\sfX$ be a measurable set, and $\text{p}(\cdot)$ be a probability density defined on $\sfX$. Let $\ev{X\sim p}{f}$ is the expectation of the function $f$ with respect to $p$. The relative entropy, or Kullback-Leibler (KL) divergence between the density $q$ and $p$ is given by $\kl{\tq}{\tp} = \ev{q}{\log q - \log p}$. The relative entropy is jointly convex in both arguments.

\textbf{Information Projection:}  Let $\cF_\sfS$ be the set of all densities supported on $\sfS \subset \sfX$. The information projection of a base density $p(\cdot)$ onto a constraint (measurable) set $\sfS \subset \sfX$ is defined as: 
\begin{equation*}
\tq_* = \argmin{\tq \in \cF_\sfS} \; \kl{\tq}{\tp}
\end{equation*}
$\cF_\sfS$ is closed and bounded for all cases of interest in this manuscript, so that $q_*$ exists.


\textbf{Submodular functions:} Let $f: \frp(d) \rightarrow \bbR$ be a set function. $f$ is a \emph{submodular} function if for all sets $\sfx, \sfy$ in its domain $f(\sfx \cup \sfy) +  f(\sfx  \cap \sfy) \leq f(\sfx)+ f(\sfy)$. Further, $f$ is \emph{normalized} if $f(\emptyset) = 0$. $f$ is monotone if for $\sfx \subset \sfy$, $f(\sfx) \leq f(\sfy)$. 
Submodular functions are of special interest because greedy algorithm and its simple variants achieve provable approximation guarantees for several otherwise NP-Hard combinatorial optimization problems~\citep{nemhauser1978, Sviridenko2004, Calinescu08}. 

\textbf{Matroids:} A matroid is a structure $(\sfN, \sfE)$, where $\sfN$ is the \emph{ground set}, and $\sfE \subset \frp(\sfN)$ is a family of \emph{independent} sets that satisfies: (i) $\sfB \in \sfE , \sfA \subset \sfB \implies \sfA \in \sfE$, and, (ii) $\sfA \in \sfE, \sfB \in \sfE, | \sfA | < |\sfB | \implies \exists x \in \sfB - \sfA  \text{ s.t. } \sfA \cup x \in \sfE$. 
%
A \emph{uniform} matroid has $\sfE$ as the set of all possible $k$ and lesser sized subsets of $\sfN$, and thus induces the $k$-cardinality constraint. Similarly, a knapsack constraint can be encoded by a matroid which has each candidate solution in $\sfE$ as a set of possible groups, each with an associated cost, such that the total cost of each candidate solution in $\sfE$ is less than or equal to the knapsack value. A \emph{partition} matroid partitions $\sfN$ into subsets $\{ \sfX_1, \sfX_2, \ldots, \sfX_r\}$, with $\sfE = \{ \sfA \, | \, \sfA \subset \sfN, | \sfA \cap \sfX_i | \leq k_i \forall i \in [r] \}$ for given $\{ k_1, k_2, \ldots, k_r\}$.

\subsection{Information Projection for Sparse Variables}
\label{sec:priors}
 
A $d$ dimensional variable $\bx$ is $k$-sparse if it is non-zero on at most $k$ dimensions. The support of the variable $\bx \in \bbR^d$ is defined as $supp(\bx):=\{i \in [d] \vert \bx_i \neq 0\}$. Similarly, a $d$ dimensional probability density $\tp$ is $k$-sparse if all random variables $\bx \sim \tp$ are $k$-sparse. Let $\sfA$ be the set of all $\frac{d!}{k!(d-k)!}$ $k$-sparse support sets. The information projection of $\tp$ onto $\cF_\sfA$ is equivalent to restriction of $\tp$ onto $\sfA$~\citep{koyejo2014}, which is a natural approach for constructing a sparse prior. Unfortunately, this information projection is generally intractable. Instead, ~\citet{koyejo2014} propose the following approximation:
\begin{equation}
\min_{\sfS \subset \sfA} \min_{\tq \in \cF_\sfS} \kl{\tq}{\tp}.
\label{eq:combinatorialKL}
\end{equation}
This information projection searches for the subset $\sfS$ that captures most of the mass of $\tp$ as measured by $\min_{\tq \in \cF_\sfS} \, \kl{\tq}{\tp}$. 
The inner optimization over $\cF_\sfS$ can be solved in closed form \citep{koyejo2014} as $\min_{\tq \in \cF_\sfS} \kl{\tq}{\tp} = - \log \tp(\bx_{\sfS^c} = 0 )$. Define the function $J: \frp(d) \rightarrow \bbR $ as $J(\sfS) := \log \tp (\bx_{\sfS^c} = 0 )$, and the function $\tilde{J}: \frp(d) \rightarrow \bbR $ as $\tilde{J}(\sfS):= J(\sfS) - J(\emptyset) $. The optimization problem~\eqref{eq:combinatorialKL} is equivalent to 
\begin{equation}
\max_{| \sfS | = k} \tilde{J}(\sfS) 
\label{eq:Js}
\end{equation}
Note that the cardinality constraint is a uniform matroid constraint. While~\eqref{eq:Js} is combinatorial, 
the following theorem ensures a good approximation by greedy support selection. 
\begin{theorem}[\citet{koyejo2014}]
$\tilde{J}(\sfS)$ is normalized monotone submodular.
\label{thm:JsSubmodular}
\end{theorem}
 Thus, a simple greedy algorithm achieves a $(1 - \frac{1}{e})$ approximate solution~\citep{nemhauser1978}. 
 

%% file: groupsparse.tex
\section{Approximate Inference via Information Projection for Structured Sparse Variables}
\label{sec:mainsparse}

In this section, we generalize the cardinality constrained information projection in two ways. First, we consider information projection subject to \emph{group} sparsity, and show that the resulting combinatorial problem of selecting the \emph{most relevant} groups is monotone submodular subject to a knapsack constraint. Secondly, we consider general matroid constraints for structured sparsity, and present an algorithm that greedily selects from the enumeration of the matroid constraint. We consider the special case of  partition matroid constraint where sets of variables are pre-grouped into \emph{views}, and seek to select variables subject to constraints on the maximum number of variables selected from each view. We leverage the research in submodular optimization to present the respective variants of the greedy algorithm that provably guarantee constant factor approximations. 

{\bf Approximate Inference:} Let $p_{0}$ be the prior distribution and $l$ be the likelihood, and $\sfA$ represent a structured subset of the domain of $p_{0}$. Restriction of the prior $p_{0}$ to the subset $\sfA$ given by $p_{\sfA, 0} \propto p_{0}(X) \indicator{X \in \sfA}$is an effective approach for constructing a prior for the structured subset $\sfA$. Unfortuntely, this restriction is generally intractable. 
We consider approximate inference by fixing $p$ as the posterior distribution $p(X) \propto p_{0}(X) l(X)$, and with $\sfA = \{\sfS_i\}$ as the set of subsets satisfying the constraint structure. e.g. for classical sparse modeling with $d$ variables, each $\sfS_i \in [d]$ is a $k$-sparse subset, and $\sfA$ is the set of all such ${d \choose k}$ subsets. In this case, the information projection to a subset $\sfS \in \sfA$ is designed to approximate Bayesian inference with respect to intractable restricted prior as $p_{\sfA}(X) \propto p_{\sfA, 0}(X) l(X)$. We note that just as in standard variational inference, the posterior projection can be implemented using $p_{0}$ and $l(\cdot)$ without explicitly computing the unrestricted posterior $p$ (useful when the $p$ is itself intractable). The proposed approach for approximate inference differs from standard approaches such as mean field variational inference, as we take advantage of the combinatorial nature of the desired structure. As such, the proposed approximate inference is most accurate when the posterior mass is well captured by the optimal subset $\sfS^* \in \sfA$. The approximate posterior is given by the information projection of $p$ onto $\cF_{\sfS^*}$, and is in fact equivalent to the projection of the restricted posterior $p_{\sfA}$ onto $\cF_{\sfS^*}$. We refer the interested reader to \citep{koyejo2014} for additional details.
\vspace{-0.1cm}
\subsection{General Structured Sparsity Constraints}
Structured sparsity extends classic sparsity constraints with additional information on the sparse subsets. For example, the sparsity could be constrained by a tree structure so that selection of a parent node implicitly selects all its children as well. The structural constraint can be encoded as a matroid ($\sfN, \sfE$) where $\sfN$ are the base set of dimensions, and $\sfE$ represents the  set of all possible candidate solutions under the given constraint. General structured sparsity is challenging to model using standard prior design techniques. Instead, one may consider Bayesian inference using the structured prior distribution recovered by restricting the base prior to the union of all possible structured subsets. As in the classic sparsity case, we consider an approximation of the resulting posterior based on the variable set which captures the maximum posterior mass. The resulting ($\sfN, \sfE$)-matroid constrained information projection of a density $p$ is simply given by:

\begin{equation}
\min_{\sfS \in \sfE} \min_{supp(\tq)\in \sfS } \kl{\tq}{\tp}
\label{eq:matroid}
\end{equation}



A simple greedy algorithm  on the enumeration of the matroid as outlined in Algorithm~\ref{algo:greedymatroid} can be used for support selection under general matroid constraints. Note that the greedy selection algorithm for the classic sparsity case~\citep{koyejo2014} is a special case of Algorithm~\ref{algo:greedymatroid} with a uniform matroid. For the more general matroid constraints, greedy selection on the enumeration admits slightly weaker guarantees. Improved approximation guarantees can be achieved by randomized algorithms~\citep{Calinescu08}. 

\begin{theorem}[\citet{Calinescu08}]
Algorithm~\ref{algo:greedymatroid} guarantees a constant factor approximation of $1/2$ for \eqref{eq:matroid}.
\end{theorem}

\paragraph{Multi view sparsity.} A special case of structured sparsity is the multi view sparsity. The base set of dimensions are divided into $v$ views/groups. Also given is a set of maximum number of allowed selections from each view $\{k_1, k_2, \ldots, k_v\}$. In other words, no more than $k_i$ selections can be made from the $i^{\text{th}}$ view/group. It should be straightforward to see that the multi view sparsity constraint induces a partition matroid structure defined in Section~\ref{sec:background}, and as such Algorithm~\ref{algo:greedymatroid} is applicable.
 Algorithm~\ref{algo:greedymatroid} can be easily re-written for the partition sparsity constraint to avoid exhaustive enumeration of the set $\sfE$ as Algorithm~\ref{algo:greedymultiview}. The $1/2$ factor approximation guarantee carries over for Algorithm~\ref{algo:greedymultiview}. We shall see in the sequel that this particular algorithm leads to an efficient inference algorithm for sparse Probabilistic CCA.


\begin{minipage}[t]{6.5cm}
\begin{algorithm}[H]
\centering
\begin{algorithmic}[1]
\STATE Input: Matroid $(\sfN, \sfE) $ 
\STATE $\sfA \leftarrow \emptyset$
\WHILE {$\sfN $ is not empty}
\STATE $s^* \leftarrow \arg\max_{s \in \sfN}  J(\sfA \cup \{s\} ) - J (\sfA)$
\IF{ $\sfA \cup\{ s^*\} \in \sfE$ }
\STATE $\sfA = \sfA \cup \{s^*\}$
\ENDIF
\STATE $\sfN = \sfN - \{s^*\}$
\ENDWHILE
\STATE \textbf{Return} $\sfA$
\end{algorithmic}
\caption{GreedyMatroid($\sfN, \sfE$)} \label{algo:greedymatroid}
\end{algorithm}

\end{minipage}
\begin{minipage}[t]{6.5cm}
\begin{algorithm}[H]
\centering
\begin{algorithmic}[1]
\STATE Input : $\sfN$, Sparsities $\{k_1, k_2, \ldots, k_v\}$ , mapping function $m : [d] \rightarrow [v]$.
\STATE $\sfA \leftarrow \emptyset$
\STATE selected[i]=0, $\forall i \in [v]$
\WHILE {$\sfN $ is not empty}
\STATE $s^* \leftarrow \arg\max_{s \in \sfN}  J(\sfA \cup \{s\} ) - J (\sfA)$
\IF{  selected[$m(s^*)$] $< k_i $ }
\STATE $\sfA = \sfA \cup \{s^*\}$
\STATE selected[$m(s^*)$] +=1 
\ENDIF
\STATE $\sfN = \sfN - \{s^*\}$
\ENDWHILE
\STATE \textbf{Return} $\sfA$
\end{algorithmic}
\caption{GreedyMultiView($k_1, k_2, \ldots, k_v$, $m(\cdot)$) \label{algo:greedymultiview}}
\end{algorithm}
\end{minipage}

\subsection{Group Sparsity Constraints}
\label{sec:groupSparse}
Group sparsity involves selecting variables from $r$ groups subject to the constraint that if a group is selected, all the variables within the group must be selected, but no more than $k$ variables can be selected in all.
Let $\sfG=\{ \sfG_1, \sfG_2, \ldots, \sfG_r\}$ represent the set of $r$ groups, so that $\forall i, \sfG_i \subset [d] $ and $\forall i \neq j, \sfG_i \cap \sfG_j = \emptyset$. As in the classic sparse case, information projection to the set of all group sparse subsets of $[d]$ is intractable in general. Instead, we propose approximate inference by seeking the projection to the set which maximizes the captured mass of $p$. The resulting group sparsity constrained information projection of a density $p$ is given by:
\begin{equation}
\min_{\sfS \subset [r]} \min_{\left\{ \tq\,|\, supp(\tq) \subset \bigcup_{i \in \sfS} \sfG_i, \, \sum_{i \in \sfS} {| \sfG_i|  \leq k} \right\}} \kl{\tq}{\tp}
\label{eq:groupsparse}
\end{equation} 

\begin{theorem}
The group selection problem~\eqref{eq:groupsparse} is equivalent to a normalized monotone submodular maximization problem with a knapsack constraint. 
\label{thm:JsGroupKnapsack}
\end{theorem}
The proof is provided in the supplement. 
We present a re-weighted greedy algorithm with partial enumeration in Algorithm~\ref{algo:groupsparsemain} to solve \eqref{eq:groupsparse}. The re-weighting is ensures that the greedy step chooses the best possible myopic marginal gain. However, with the re-weighting alone the approximation factor can be arbitrarily bad. To bound it to a constant factor, partial enumeration is required. We also note that Algorithm~\ref{algo:groupsparsemain} is \emph{not} a special case of Algorithm~\ref{algo:greedymatroid}, as it exploits the special structure of group sparsity to construct a scheme with improved optimization-theoretic guarantees. The following theorem establishes the optimization guarantee of Algorithm~\ref{algo:groupsparsemain}.
\begin{theorem}[\citet{Sviridenko2004}] Algorithm~\ref{algo:groupsparsemain} with $m=3$ guarantees a constant factor approximation of $(1 - \frac{1}{e})$ for \eqref{eq:groupsparse}.
\end{theorem}

\begin{minipage}[t]{6.5cm}
\begin{algorithm}[H]
\centering
\begin{algorithmic}[1]
\STATE Input: Set of groups $\sfG$, Total max sparsity $k$, parameter $m$, cost function $c(\cdot)$
\STATE $\sfS_1 \leftarrow \arg\max_{\sfs \subset \sfG,  |\sfs| < m, c(\sfs) \leq k}  \tilde{J}(\sfs)$
\STATE $\sfS_2 \leftarrow \emptyset$
\FORALL{   $\sfs \subset \sfG, |\sfs| = m,  c(\sfs) \leq k $ }
\STATE $\sfS_3 \leftarrow $ ReweightedGreedy($\sfG$, $k - m-1$, $c(\cdot)$, $\sfs$)
\IF {$\tilde{J}(\sfS_2) \leq \tilde{J}(\sfS_3) $}
\STATE $\sfS_2 \leftarrow \sfS_3$
\ENDIF
\ENDFOR
\STATE \textbf{Return} $\arg\max\{\tilde{J}(\sfS_1), \tilde{J}(\sfS_2)\}$
\end{algorithmic}
\caption{GreedyPartialEnum ($\sfG, k, c(\cdot)$) \label{algo:groupsparsemain}}
\end{algorithm}

\end{minipage}
\hspace{0.5cm}
\begin{minipage}[t]{7cm}
\begin{algorithm}[H]
\centering
\begin{algorithmic}[1]
\STATE Input: Set of groups $\bar{\sfG}$, Total max sparsity $\bar{k}$, cost function $c(\cdot)$, Init groups $\bar{\sfS}_2$
\STATE $\sfA \leftarrow  \bar{\sfS}_2$
\WHILE{ $\bar{\sfG}\backslash\sfA \neq \emptyset$  }
\STATE  $ \sfs^* \leftarrow \max_{\sfs \in \bar{\sfG} \backslash \sfA} \frac{ J(\sfA \cup \sfs) - J (\sfA)}{ c(\sfs) }$
\IF{ $c(\sfA \cup \sfs^*) \leq k$}
\STATE $\sfA = \sfA \cup \sfs^*$
\ENDIF
\STATE $\bar{\sfG} = \bar{\sfG} - \sfs^*$
\ENDWHILE
\STATE \textbf{Return $\sfA$}
\end{algorithmic}
\caption{ReweightedGreedy ($\bar{\sfG}$, $\bar{k}$,$c(\cdot)$,  $\bar{\sfS}_2$)}
\label{algo:reweighted}
\end{algorithm}
\end{minipage}

%% file: applications.tex
\section{Applications: Probabilistic Models with Matroid Constrained Variables}
\label{sec:apps}
While the class of probabilistic models that admit a representation via matroid constraints is quite broad, we consider special cases in detail (i) group sparse principal components analysis, (ii) sparse canonical correlation analysis, and (iii) group sparse regression (in the supplement). 
The framework developed in Section 3 readily yields efficient greedy solutions for feature selection for all three cases.

\subsection{Group Sparse Probabilistic Principal Components Analysis}
\label{sec:pca}

Probabilistic PCA aims to factorize a matrix $\bT \in \bbR^{n \times} $ as $\bT \approx \bx \bw^\top$, where $\bx \in \bbR^n$ is a deterministic vector, and $\bw \in \bbR^d$ is a random variable. For simplicity, we only consider the rank 1 case i.e. where $\bx, \bw$ are vectors. The general matrix case follows using standard deflation techniques for multiple factors~\citep{khanna2015, tipping1999, khanna2015deflation}. The generative model for the observed data matrix is $\bT = \bx \bw^\top + \epsilon$, where $\epsilon \sim \cN(0, \sigma^2)$. We consider the case where the prior $\bw \sim \cN(\mathbf{0}, \bC)$, and in addition, $\bw$ is assumed be sparse. Let $\theta = \{\bx, \sigma\}$ represent the set of deterministic parameters.

The underlying $\bx, \bw$ may be estimated by maximizing the log likelihood using Expectation Maximization (EM), which optimizes for $\bx$ and $\bw$ in an alternating manner in the M-step and the E-step respectively. The algorithm can be interpreted as minimizing \emph{free energy} cost function~\cite{neal98} given by:
\begin{equation*}
\label{eqn:energyFunction}
\mathscr{F}(\tq(\bw),\theta) = -\kl{ \tq(\bw)}{ \tp(\bw |
\bT; \theta)} +\log\tp(\mt;\theta),
\end{equation*}
where $\log\tp(\mt;\theta)$ is the marginal log-likelihood.
The M-step is a search over the parameter space, keeping the latent random variable $\bw$ fixed. Similarly, the E-step is the search over the space of distribution $\tq$ of the latent variables $\bw$, keeping the parameters $\theta$ fixed 
\begin{equation*}
\text{M-step: } \max_{\theta} \mathscr{F}(\tq(\bw),\theta), 
\quad
\text{E-step: } \max_{\tiny{ \tq}} \mathscr{F}(\tq(\bw),\theta).
\end{equation*}
This view of the EM algorithm provides the flexibility to design algorithms
with any E and M steps that monotonically increase $\mathscr{F}$. When the search space of q in the E-step is unconstrained, E-step outputs the posterior p($\bw | \bT; \theta $). 
Constraining the search space of q leads to a \emph{variational E-step}. In this section, we consider a restriction which approximates the combinatorial space of group sparse distributions using the framework developed in Section~\ref{sec:groupSparse}.

We now derive the explicit equations to apply Algorithm~\ref{algo:groupsparsemain}. 
The posterior p($\bw | \bT; \theta $) is Gaussian with p $\sim \cN(\bmu, \bSigma)$, where $\bSigma^{-1} = \bC^{-1} + \frac{\| \bx\|^2_2}{\sigma^2}$, and $\bmu = \frac{1}{\sigma^2}\bSigma \bT^\top \bx$. Define $\br:= \bSigma^{-1} \bmu$. Expanding the KL divergence for information projection from~\eqref{eq:groupsparse}, simple algebra yields that the support selection requires the following submodular maximization problem: 
\begin{equation*}
\max_{\left\{\sfS \subset [r],\, \sfS = \bigcup_{i \in \sfS} \sfG_i,\, |\sfS| \leq k \right\}} \br_\sfs^\top [\bSigma^{-1}]_\sfs\br_\sfs -  \log\det [\bSigma^{-1}]_\sfs.
\end{equation*}
The resulting approximate posterior is given by the respective conditional $\tq^*(\bw) = \tp(\bw | \bw_{\sfS^c} =0)$ (c.f. \citep{khanna2015}). 
The M-step equations for $\bx$ and $\sigma^2$ are also easily obtained as closed form updates(c.f. \citep{khanna2015}).

\subsection{Sparse Probabilistic Canonical Correlation Analysis (Sparse PCCA)}
\label{sec:cca}
%
%

Probabilistic CCA~\cite{Bach05pcca, klami2007, archambeau2008} is a multi-view generalization of Probabilistic PCA
i.e.. multiple \emph{views} of the data are observed.
Hence, we observe $n$ samples of $d_1, d_2, \ldots, d_v$ as matrices $\bT_1, \bT_2, \ldots, \bT_v$ each of which are one of the $v$ views of the observed. 
The generative model assumes an underlying parameter $\bx \in \bbR^n$ shared among all the views, and the random variables $\{\bw_i \in \bbR^{d_i}, \forall i \in [v] \}$. As in Section~\ref{sec:pca}, we note that $\bx, \{\bw_i\}$ can be matrices in general. For clarity, we focus on modeling for the top-1 component. The framework is easily extended to multiple factors using deflation techniques.
The random variables are drawn from Gaussian distributions $ \bw_i \sim \cN(\mathbf{0}, \bC_i), \, \forall i \in [v]$, and each of the view is generated as  $\bT_i = \bx \bw^\top + \epsilon, \, \forall i \in [v]$, where the noise $\epsilon \sim \cN(0, \sigma^2)$. Further, we wish to infer sparse $\bw_i$ so that $\forall i \in [v], | \text{supp}(\bw_i) | \leq k_i$ for the supplied $k_i$. 
The parameters are optimized using an EM algorithm. The variational E-step can be formulated to honor the sparsity constraints on the random variables. We next that show that the variational E-step solves a submodular maximization problem subject to a partition matroidal constraint.


We now map the sparse PCCA problem to the partition matroidal constrained optimization. Let  $\bT = [\bT_1, \bT_2, \ldots, \bT_v ]$ be the matrix of size $n \times (\sum_i d_i)$ constructed by stacking all the observed views column-wise. Similarly, $\bw = [ \bw_1; \bw_2; \ldots; \bw_v]$ be the vector obtained by end-to-end concatenation of random variable vectors of all views. Define $\bC \in \bbR^{ (\sum_i d_i) \times  (\sum_i d_i)}$ as the block diagonal matrix with $\bC_i$ as its block. The generative model of PCCA can now be equivalently and succinctly encoded as $\bT = \bx \bw^\top + \epsilon$ where $\bw \sim \cN(\mathbf{0}, \bC)$, and, $\epsilon \sim \cN(0,\sigma^2)$.  Further, the partition matroid is easy to construct with $\sfN = [ \sum_i d_i]$, and $\sfA_i$ to to be the respective index set of $\bw_i$ in $\bw$. Again, proceeding as in Section~\ref{sec:pca}, the submodular maximization problem can be written as: 
\begin{equation*}
\max_{\left\{\sfS \in \sfE,\, \text{Matroid}(\sfN,\sfE) \right\}} \br_\sfs^\top [\bSigma^{-1}]_\sfs\br_\sfs -  \log\det [\bSigma^{-1}]_\sfs.
\end{equation*}
Hence, Algorithm~\ref{algo:greedymatroid} or equivalently Algorithm~\ref{algo:greedymultiview} can be used for sparse inference. We focus on sparse CCA for this manuscript. However, it should be easy to the see that further extension to group sparse CCA is straightforward by modifying the constraining partition matroid appropriately.

%% file: experiments.tex
\section{Experiments}
\label{sec:experiments}
We now present empirical results comparing the proposed information projection based support selection technique to established baselines for 2 applications for real world datasets, namely group sparse PCA, and sparse CCA. For model verification, additional experiments on simulated data for group regression are presented in the supplement. We implement our method in Python using Numpy and Scipy libraries. The greedy selection is parallelized by Message Passing Interface using \texttt{mpi4py}. We make use of Woodbury matrix inversion identity in the cost function to greedily build up the cost function. This avoids taking explicit inverses that can lead to inconsistencies.

\subsection{fMRI data}
\textbf{Neurovault data}
A key question in functional neuroimaging is the extent to which task brain measurements incorporate distributed regions in the brain. One way to tackle this hypothesis is to decompose a collection of task statistical maps and examine the shared factors.~\citet{smith2009correspondence} considered a similar question using the brain map database decomposed via ICA, showing correspondence between task activation factors and resting state factors. Following their approach, we downloaded 1669 fMRI task statistical maps from neurovault\footnote{http://neurovault.org/}. Each image in the collection represents a standardized statistical map of univariate brain voxel activation in response to an experimental manipulation. The statistical maps were downsampled from $2mm^3$ voxels to $3mm^3$ voxels using the nilearn python package\footnote{{http://nilearn.github.io/}}. We then applied the standard brain mask, removing voxels outsize of the grey matter, resulting in $d$=65598 variables. We incorporate smoothness via spatial precision matrix $\bC^{-1}$ on the prior on $\bW$ which is generated by using the adjacency matrix of the three dimensional brain image voxels. This directly corresponds to the observation that nearby voxels tend to have similar functional behavior.

\begin{figure}\centering
\begin{subfigure}[b]{.45\columnwidth}
\includegraphics[scale=0.3]{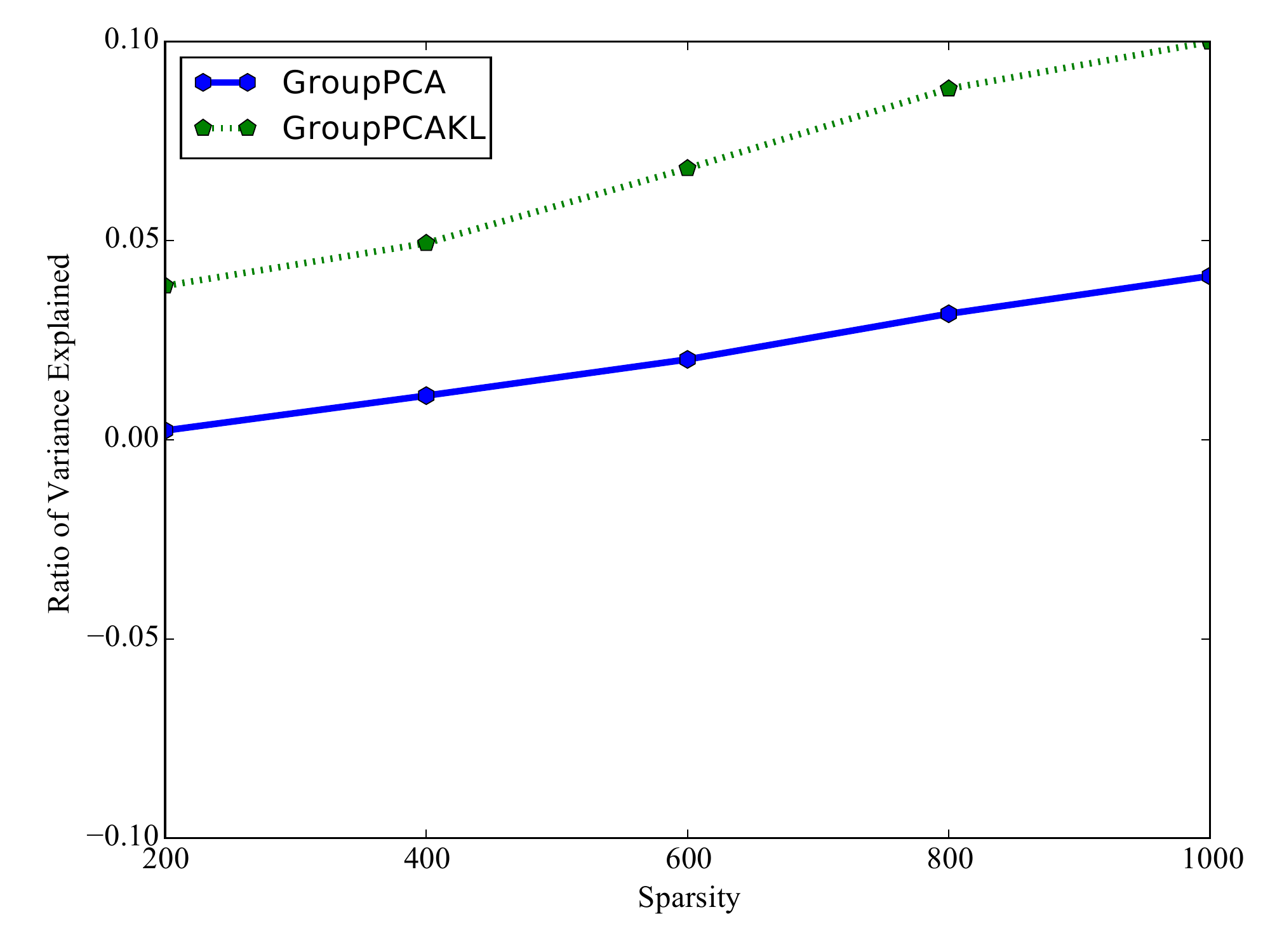}
\subcaption{Group Sparse PCA performance on the Neurosynth data}
\label{fig:pca}
\end{subfigure}
\hspace{0.5cm}
\begin{subfigure}[b]{.45\columnwidth}
\includegraphics[scale=0.3]{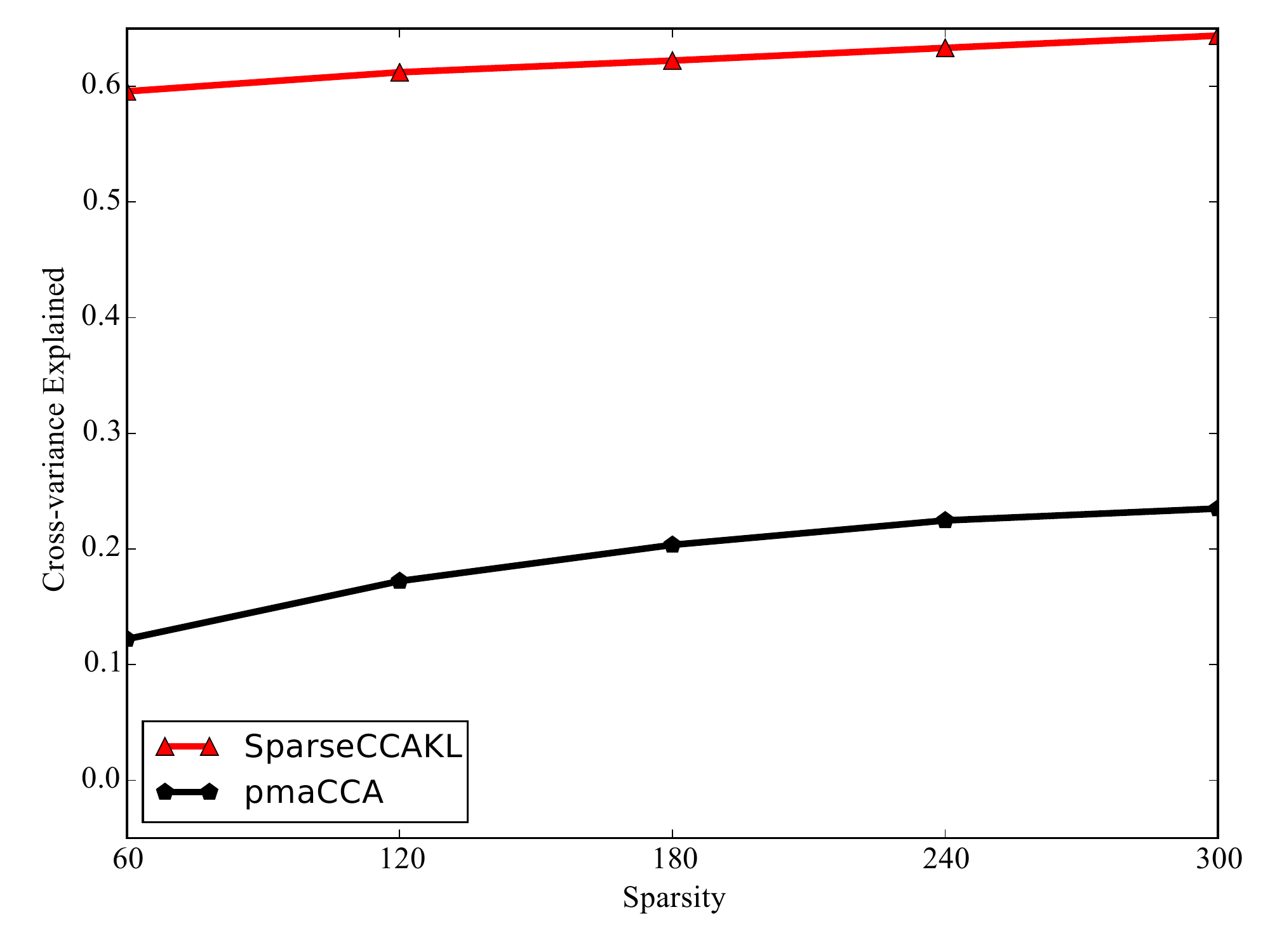}
\caption{Sparse CCA performance on n-back task Human Connectome Project data}
\label{fig:cca}
\end{subfigure}
\end{figure}

While our greedy algorithm can easily scale to dimensionality of size 65598, the matlab implementation of the baseline is not as scalable. We cluster the original set of dimensions to $d=$10000 dimensions using the spatially constrained Ward hierarchical clustering approach of \citet{michel2012}. We further apply the same hierarchical clustering to group the dimensions into 500 groups, with group sizes ranging from 1 to 1500 with average group size close to 20. We apply our information projection based Group Sparse PCA algorithm (GroupPCAKL) developed in Section~\ref{sec:pca}. The group sparse constraint specifies that each group can be either wholly included or completely discarded from the model. Our algorithm adheres to this specification. It is possible to have a soft version of the constraint which allows for sparsity within each chosen group. This is typically imposed as a regularization trade-off between sparsity across and within groups. We compare against the Structured Sparse  PCA algorithm (GroupPCA) of~\citet{jenatton2010sspca}, which is considered state of the art algorithm for group sparse PCA. We report the ratio of variance explained by the top $k$-sparse eigenvector at different values of $k$ and show superior performance of GroupPCAKL in Figure~\ref{fig:pca}.

\textbf{Human Connectome Project}
Another interesting question that the neuroscientists are interested to address is about the association of human brain function to human behavior. The brain function and the human behavior can be thought of as two \emph{views} of underlying latent traits. This intuition suggests possible application of the CCA based approaches (Section~\ref{sec:cca}). We make use of the Human Connectome Project data (HCP)~\citep{VanEssen201362} for this purpose. It consists of large number of samples of high quality brain imaging and behavioral information collected from several healthy adults. We specifically use two datasets of different tasks - 2K (2 Back vs 0 Back contrast, measures working memory), and REL-match (REL vs MATCH contrast, measures relational processing)\footnote{https://wiki.humanconnectome.org/display/PublicData/Task+fMRI+Contrasts}. We download and extract brain statistical maps (a statistical map is a summary of each voxel in the brain in response to externally applied controlled stimulus) and respective behavioral variances from 497 adult subjects. Each subject has 380 behavioral variables, 27000 downsampled voxels. Further details on the task are available in the HCP documentation~\citep{VanEssen201362}. On the extracted maps, we perform the standard preprocessing for motion correction, and image registration to the MNI template for consistency of comparisons across subjects.  The resulting maps we downsampled in the similar way as the Neurosynth data. 

As before, to incorporate smoothness we use the spatial correlation matrix as the prior on the factors of view of statistical map. For the view of behavioral data, we use an identity matrix as the respective prior covariance matrix. We apply our Information Projection based Sparse CCA (SparseCCAKL) approach and compare it against the Sparse CCA algorithm developed by \citet{Witten2009} (pmdCCA) which is used in its original or slightly modified form as state of the art in many newuroscience and biomedical applications. For quantitative comparison, we use the n-back task dataset to report the cross-variance explained which is defined as follows. If $\bX, \bY$ are the two views, and $\bu,\bv$ are respective CCA (possible sparse) factors, the cross-variance is defined as :$\frac{\bu^\top \bX^\top\bY \bv}{| \bu^\top\bX\bu ||\bv^\top\bY\bv|}$. We show strong performance of SparseCCAKL on the metric in Figure~\ref{fig:cca}.

\begin{figure}
\centering
\includegraphics[width=1.0\linewidth]{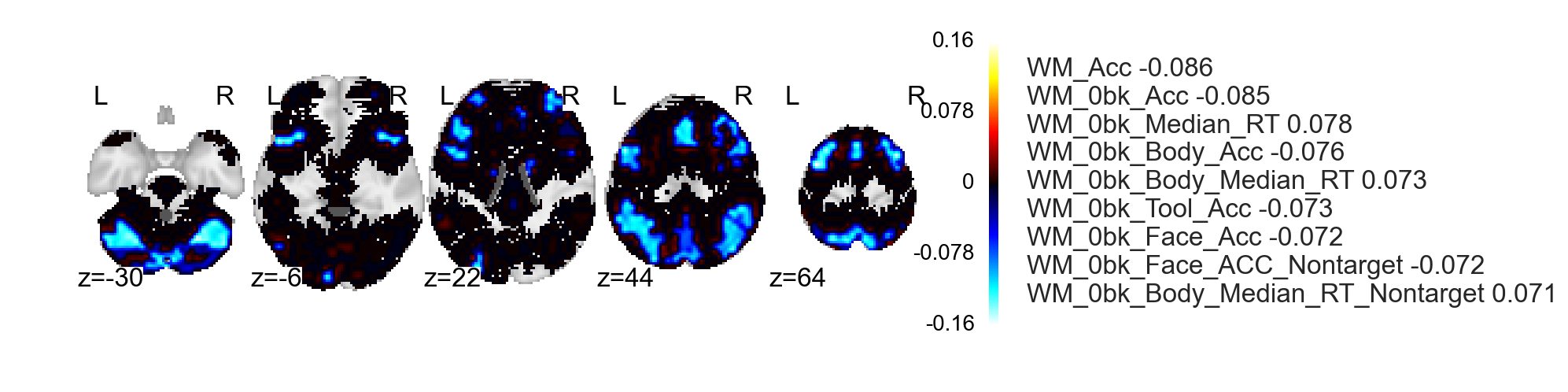}
\caption{The first factor from 2-back task. Neural support is seen in a number of frontal and parietal regions and cerebellum, consistent with cognitive control systems usually engaged by the task.  Behavioral correlates including both reaction time and accuracy on the task, showing greater neural engagement associated with slower and less accurate performance.}
\label{fig:cca_2k}
\end{figure}

\begin{figure}
\centering
\includegraphics[width=1.0\linewidth]{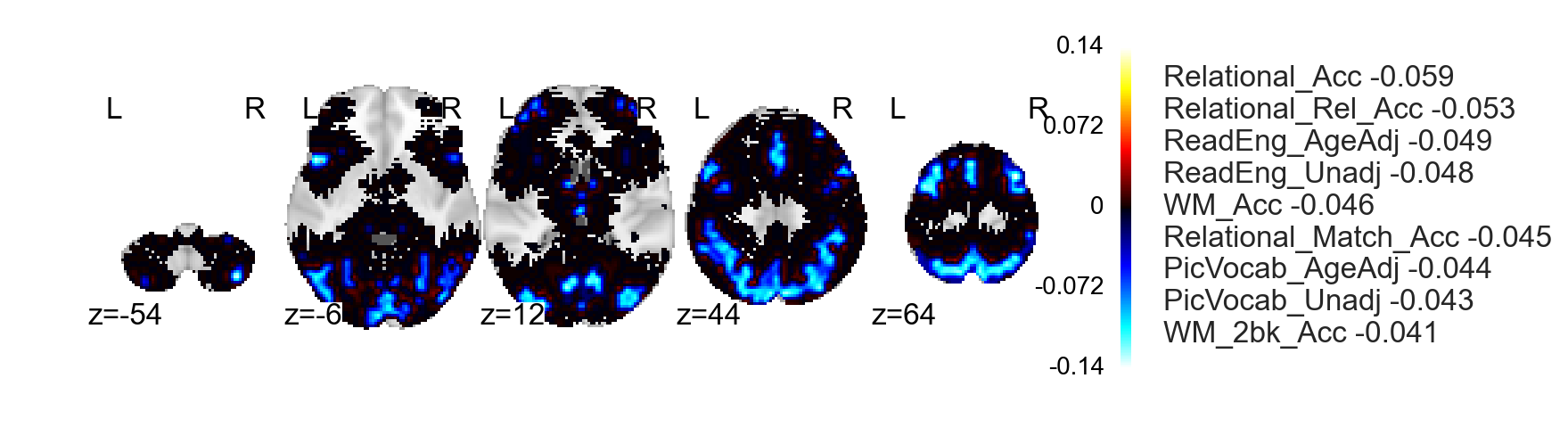}
\caption{The first factor from relational reasoning task. Neural support is observed in frontal, parietal, and occipital cortex.  Behavioral correlates captured both performance on this particular task, as well as independent measures related to higher cognitive functions including working memory capacity, vocabulary, and reading.}
\label{fig:cca_REL}
\end{figure}

%% file: conclusion.tex
\section{Conclusion and Future Work}
\label{sec:conclusion}
This manuscript proposes efficient algorithms for approximate inference via information projection that are applicable to any structure on the set of variables which admits enumeration using a matroid. The class of probabilistic models that can be expressed in this way is quite broad. In particular, we highlight the special cases of group sparse regression, group sparse principal components analysis and sparse canonical correlation analysis. We also presented empirical evidence of strong performance compared to established baselines of respective models on simulated and two real world fMRI datasets. Our strong results motivates us to  further study the theoretical properties of the information projection framework, including sparsistency and robustness. 

%% file: appendix.tex
\section{Proof of Theorem~\ref{thm:JsGroupKnapsack}}

\begin{proof}
We prove by mapping~\eqref{eq:groupsparse} to an equivalent problem by performing a variable change. 

Let $ \sfG_\sfS := \bigcup_{i \in \sfS} \sfG_i$. Note that the inner optimization $\min_{| \sfG_\sfS | \leq k, q \in \cF_{\sfG_\sfS}} \kl{\tq}{\tp} = - \log p( \bx_{\sfG \backslash \sfG_\sfS}) $~\citep{koyejo2014}. 

Define the function $J: \frp(r) \rightarrow \bbR $ as $J(\sfS) := \log \tp (\bx_{\sfG \backslash \sfG_\sfS} = 0 )$, and the function $\tilde{J}: \frp(r) \rightarrow \bbR $ as $\tilde{J}(\sfS):= J(\sfS) - J(\emptyset) $. 

Define the costs associated with picking $\sfG_i$ as $c_i = | \sfG_i | \forall i \in [r]$. The cost function of a set $\sfs \subset \sfG$ can thus be written as $c(\sfs) := \sum_{\forall i \text{ s.t. } \sfG_i \in \sfs} c_i$
The optimization problem~\ref{eq:groupsparse} is then equivalent to 
$\max_{   \sum_{i \in \sfS}  c_i \leq k} \tilde{J}(\sfS) $. 

The result follows from Theorem~\ref{thm:JsSubmodular}.

\end{proof}

\section{Application: Group Sparse Linear Regression}
\label{sec:regression}
Consider a generative model for $n$ samples given by a linear model and an additive Gaussian noise: $\by= \bZ \bbeta + \epsilon,$, where $y \in \bbR^n$ is the response, $\bZ \in \bbR^{n \times d}$ is the feature matrix, and $\bbeta \in \bbR^d$ is the vector of regression weights.  The weights have an associated normal prior, $\bbeta \sim \cN(\mathbf{0}, \bC)$ for a known $\bC$. The noise $\epsilon$ is drawn from a Gaussian $\epsilon \sim \cN(0, \sigma^2)$. The posterior distribution of $\beta$ is also a Gaussian, $\tp(\bbeta | \by ) \sim \cN(\bmu, \bSigma)$ and can be written in closed form by standard Bayes theorem with $\bSigma^{-1} =  \bC^{-1} + \frac{1}{\sigma^2} \bZ^\top\bZ, $ and, $\bmu = \frac{1}{\sigma^2}\bSigma\bZ^\top\by$.

Let $\sfG = \{\sfG_1, \sfG_2, \ldots, \sfG_r\}$ be the given set of groups so that $\forall i \in [r], \sfG_i \subset [d],\,$ and $ \forall i \neq j,  \sfG_i \cap \sfG_j = \emptyset $. The optimization problem for sparse group selection is then given by~\eqref{eq:groupsparse}. For the spacial case where $\tp$ is Gaussian, the information projection to any structured subset remains in the Gaussian family~\citep{koyejo2013}. Thus, the search for q in~\eqref{eq:groupsparse} can be restricted to Gaussians.
Define $\br = \frac{1}{\sigma^2} \bZ^\top \by$. It is easy to show by expanding the KL that~\eqref{eq:groupsparse} for group sparse linear regression is equivalent to the submodular maximization problem: 
\begin{equation}
\max_{\left\{ \sfS \subset [r],\; \sfS = \bigcup_{i \in \sfS} \sfG_i, \; |\sfS| \leq k \right\}} \br_\sfs^\top [\bSigma^{-1}]_\sfs\br_\sfs -  \log\det [\bSigma^{-1}]_\sfs.
\label{eq:regression}
\end{equation}
Once the support $\sfs$ is selected, the respective approximate posterior $\tq^*$ can be obtained as the respective conditional $\tq^*(\bx) = \tp(\bx | \bx_{\sfS^c} =0)$.
\subsection{Experiment: Simulated data}
\begin{figure}[th]
\centering
\begin{subfigure}[b]{.45\columnwidth}
\includegraphics[scale=0.31]{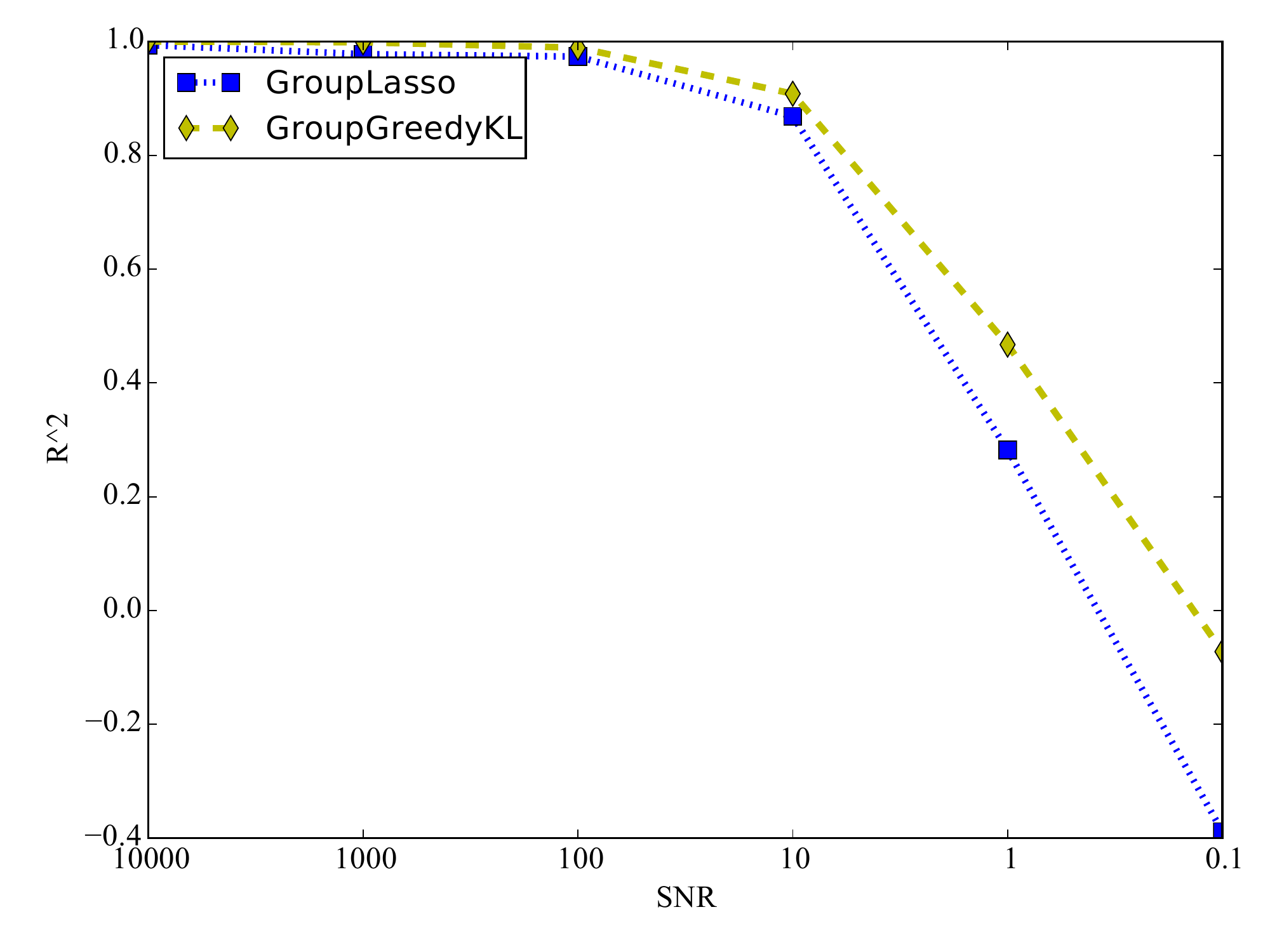}
\subcaption{$R^2$ performance on test data}
\end{subfigure}
\begin{subfigure}[b]{.45\columnwidth}
\includegraphics[scale=0.31]{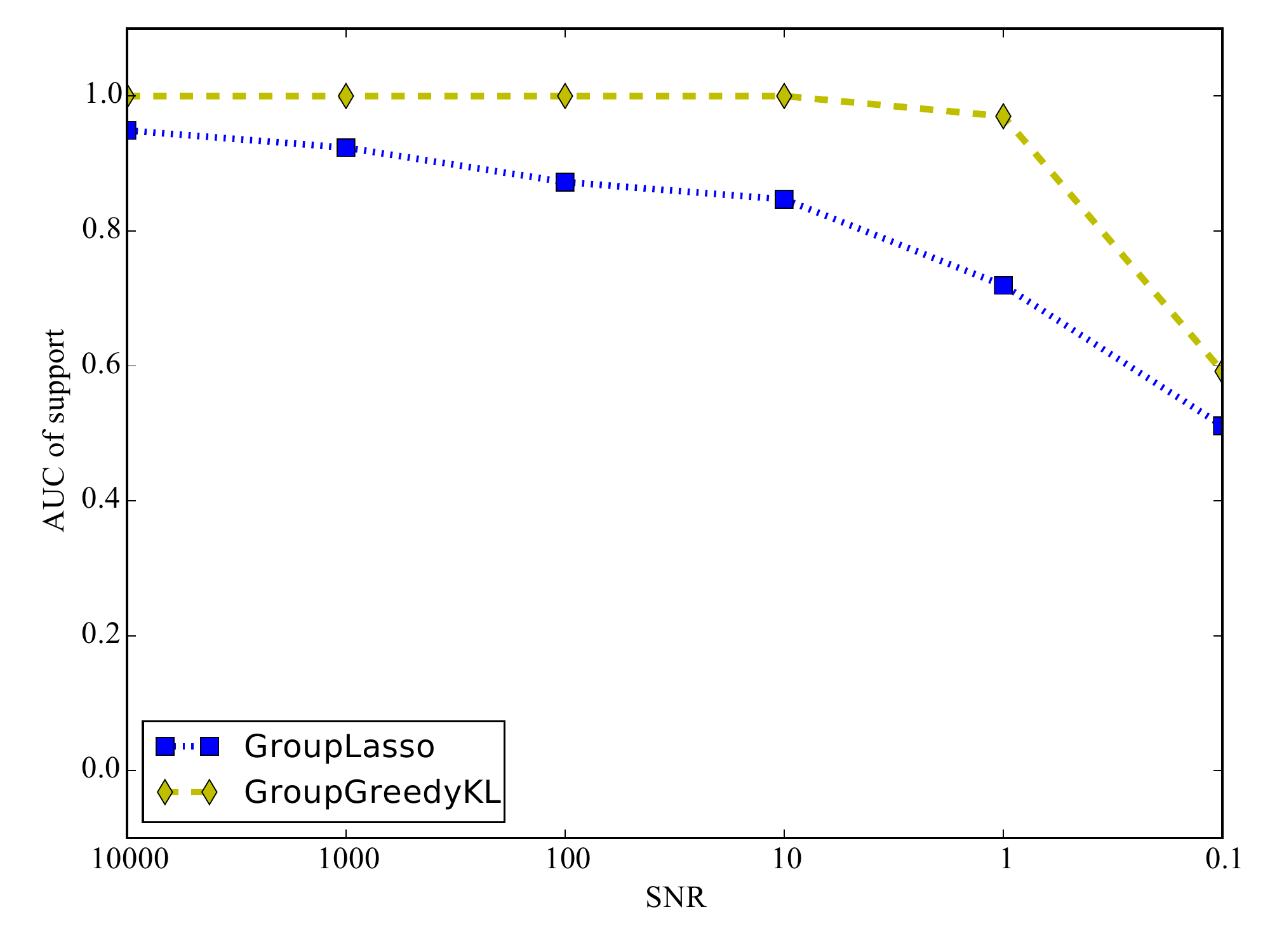}
\subcaption{Recovery of true support}
\end{subfigure}
\caption{Group Sparse Regression performance on simulated data.}
\label{fig:simulated}
\end{figure}

We compare the proposed approach for group sparsity sparsity against the sparse-group lasso~\citep{Simon13asparse-group} implemented in the package SLEP~\citep{Liu10slep} which is used in practice as state of the art.
We fix the ambient dimension to be $d=1000$. We generate an arbitrary fixed weight vector $\bbeta \in \bbR^d$ with all but $k=20$ dimensions zeroed out, arbitrarily separated into 5 groups of 4 each. We sample from the $d$-variate normal distribution with identity covariance $n=1000$ times to get the feature matrix $\bX \in \bbR^{n \times d}$. Finally we obtain the response vector $\by = \bX\bbeta + \epsilon $, where $\epsilon \sim \cN(0, \sigma^2)$ with $\sigma^2$ being set with varying values of the Signal-to-Noise ratio (SNR) so that SNR=$\{10000,1000,100,10,1,0.1\}$ to generate 6 datasets. Note that SNR $< 1$ implies variance of the noise is more than that of the signal. We split the data $50-10-40$ into training, validation and test sets. We compare performance of GroupGreedyKL (group selection based on KL projection) and GroupLasso~\citep{Simon13asparse-group} on two metrics - the AUC of the support recovered, and $R^2$ on test data. We use Bayes Factor to estimate $k$ for GroupGreedyKL. For GroupLasso, we do a parameter sweep to get the best performing numbers. For each of the 6 different SNRs, data is generated 10 different times randomly and the average results are reported. The results are presented in Figure~\ref{fig:simulated}. GroupGreedyKL performs consistently better than GroupLasso, and degrades more gracefully as SNR decreases.

